\documentclass{article} 
\usepackage{iclr2021_conference,times}
\usepackage{graphicx}
\usepackage{wrapfig}
\usepackage{tikz}
\usepackage{pgfplots}
\usepackage{amssymb}
\usepackage{amsmath}
\usepackage{amsthm}
\usepackage{pdfpages}
\usepackage{adjustbox}
\usepackage{blindtext}
\usepackage{rotating}
\usepackage[linewidth=1pt]{mdframed}
\usetikzlibrary{datavisualization}
\usetikzlibrary{arrows.meta}
\usetikzlibrary{shapes}
\usetikzlibrary{plotmarks}
\usepackage{array}

\usepackage{amsmath,amsfonts,bm}
\usepackage{subcaption}

\def\eqref#1{equation~\ref{#1}}

\def\1{\bm{1}}

\DeclareMathAlphabet{\mathsfit}{\encodingdefault}{\sfdefault}{m}{sl}
\SetMathAlphabet{\mathsfit}{bold}{\encodingdefault}{\sfdefault}{bx}{n}

\usepackage{hyperref}
\usepackage{url}

\title{Difference-in-Differences: Bridging Normalization and Disentanglement in PG-GAN}

\author{Xiao Liu$^*$, Jiajie Zhang$^*$, Siting Li\thanks{Equal contributions.}\ , Zuotong Wu, Yang Yu \\
Tsinghua University\\
\texttt{\{liuxiao17, jiajie-z19, li-st19, wuzt19\}@mails.tsinghua.edu.cn} \\
\texttt{yangyu1@tsinghua.edu.cn}
}

\newcommand{\xiao}[1]{\textbf{\color{purple}[(Xiao: #1)]}}
\iclrfinalcopy 

\newcommand{\beq}[1]{\vspace{-0.1in}\begin{equation}#1\end{equation}\vspace{-0.1in}}
\newcommand{\vpara}[1]{\vspace{0.07in}\noindent\textbf{#1}\xspace}
\newcommand{\hide}[1]{}

\newtheorem{myDef}{Definition}
\newtheorem*{citeDef}{Literature Definition}

\begin{document}

\maketitle

\begin{abstract}
What mechanisms causes GAN's entanglement? Although developing disentangled GAN has attracted sufficient attention, it is unclear how entanglement is originated by GAN transformation. We in this research propose a difference-in-difference (DID) counterfactual framework to design experiments for analyzing the entanglement mechanism in on of the Progressive-growing GAN  (PG-GAN). Our experiment clarify the mechanisms how pixel normalization causes PG-GAN entanglement during a  input-unit-ablation transformation. We discover that pixel normalization causes object entanglement by in-painting the area occupied by ablated objects. We also discover the unit-object relation determines whether and how pixel normalization causes objects entanglement. Our DID framework theoretically guarantees that the mechanisms that we discover is solid, explainable and comprehensively.

\end{abstract}

\section{Introduction}

\begin{sidewaysfigure}
\includegraphics[width=\columnwidth]{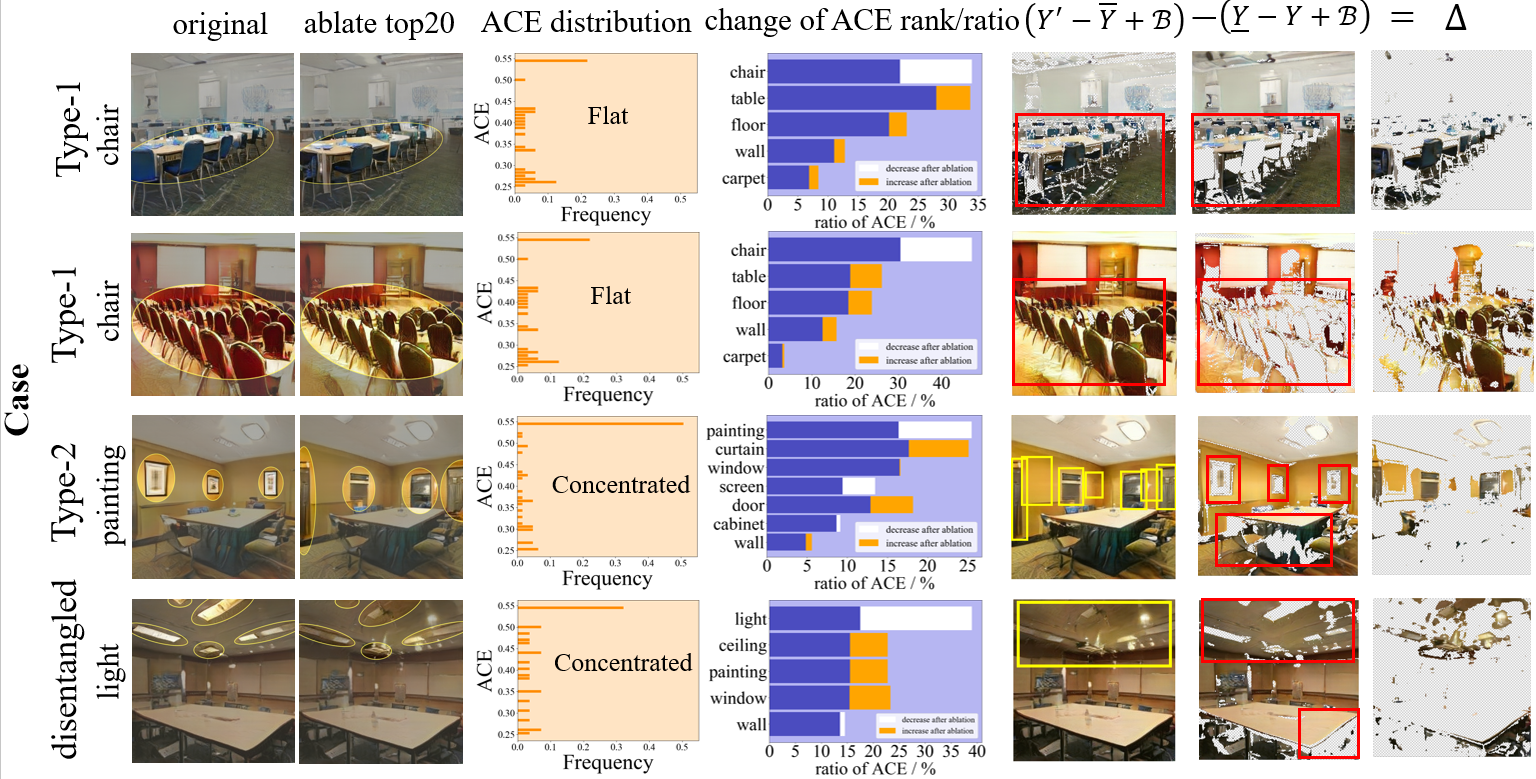} \label{fig:main_figure}
\caption{Column 1 are the original figures. Column 2 are the figures with top20 target-object-ACE units ablated. Column 3 represents the ACE-distribution of target-object-related units, where chair(Type-1 entangled) has flat ACE-distribution and is difficult to remove, while painting(Type-2 entangled) and light(disentangled)  has concentrated ACE-distribution thus are easy to remove. Column 4 shows the change of ratio and rank of ACE for different objects after ablate top20 target-object-ACE units. The increase of ACE for some objects after ablating painting-ACE units is so large that they emerge after paintings are removed.  Column 5 presents how ablated figure is changed by adding the top-20 target-object-ACE units. Where red frames and blank mean disappearance while yellow frames mean appearance. Column 6 presents how original figure is changed by ablating the top-20 ACE units back when pixel normalization coefficient is fixed to before-ablation level. Column 7, which is the difference between column 5 and column 6, presents the effect of pixel normalization: preservation and substitution.}
\end{sidewaysfigure}

Understanding the entanglement stands on the stage center of the deep learning research because entanglement is deeply rooted in the complex computational process of neural network models \citep{karpathy2015visualizing,kulkarni2015deep,higgins2016beta} while indicating non-predicable biases. Therefore, developing an output-disentangle neural network model has attracted a significant amount of attention from deep-learning society. However, the absence of analytical understanding about the mechanism causing output entanglement prevents us from discussing whether and when a neural network’s architecture can systematically avoid relative biases.

On the other hand, it is a challenge to examine the mechanism causing GAN \citep{goodfellow2014generative,radford2015unsupervised,zhang2019self,chen2016infogan} entanglement. GAN’s deep neural network structure obstructs the progress of theoretical analyses, while the experimental approach proposed by the most recent studies \citep{zhou2018interpretable,selvaraju2017grad,simonyan2013deep,olah2018building,schwab2019cxplain} is incapable of enlightening GAN’s inside structure. Current experimental approaches in the literature are designed for generating counterfactual scenarios \citep{imbens2015causal,pearl2009causality} with and without the input changes \citep{bau2017network,represent}. In contrast, understanding GAN’s internal mechanism causing entanglement asks for an experimental design that can generate counterfactual scenarios with and without GAN’s functioning. Thus, a new experimental design is necessary for studying the mechanism of GAN entanglement. 

We in this research develop a difference-in-difference (DID) \citep{2018Dif,goodman2018difference,abadie2005semiparametric,athey2006identification} experiment to analyze the entanglement mechanism originated in the pixel-normalization operation of the Progressive-growing generative adversarial network (PG-GAN) \citep{karras2017progressive}. We select to research PG-GAN because of two reasons. First, PG-GAN is an approach of generating a high-resolution figure including various objects that can entangle with each other. Second, the recent progress in literature has well prepared for applying DID to study PG-GAN’s entanglement. (\cite{besserve2018counterfactuals}) rigorously defined the concept of operation-based object disentanglement in a figure generated by GAN. (\cite{bau2018gan}) has developed an approach to clarify the causal ties between input units and output objects. Based on these two studies, we design a DID experiment to examine how pixel normalization causes object entanglement during a unit-ablating transformation. 

Our results conclude that pixel normalization causes entanglement by in-painting the area belonging to ablated objects. Once the in-painted objects are different from those surrounding the in-painted area, an entanglement effect occurs. Entanglement caused by pixel normalization can be further clustered according to the types of in-painted objects. If the in-painted objects are from the same types as the ablated one, the entanglement effect deactivates the ablating transformation. We refer to this type of entanglement as the “deactivating-ablation entanglement" or Type-1 entanglement. Otherwise, the entanglement effect causes unwanted objects’ appearance associated with ablating, referred to as the “mis-ablation entanglement” or Type-2 entanglement by this research. The difference-in-difference experiment also clarifies that the characteristics of unit-object causal relation determine the types of in-painted objects. 

We summarize our three contributions in this research:
\begin{itemize}
    \item We explain the internal mechanism of how pixel normalization causes object entanglement under ablating transformation. Because of our insights into the black-box of PG-GAN, we also clarify the necessary conditions when objects are disentangled.
    \item We clarify the mechanism how unit-object causal relation determine whether and how disentanglement occurs. 
    \item  We propose an experimental approach to analyze PG-GAN’s functioning mechanism based on the DID counterfactual framework, which can be generalized for broader deep neural network studies. Designing appropriate DID experiments to examine the functioning mechanism of neural network deserves further discussion.
\end{itemize}
Our understanding about entanglement provide a new perspective on entanglement research, which can enable of a sequence further research. For instance, it is possible to design an ablation method rather than modifying PG-GAN to avoid entanglement according to the understanding about how pixel normalization causes entanglement over objects. The understanding the deactivating-ablation entanglement also allows us to examine the robustness of objects in a figure once some input units are unexpected losses.

\section{Preliminary: PG-GAN and Disentanglement} \label{sec:pg-gan}
\begin{wrapfigure}{r}{5cm}
\centering
\includegraphics[width=2in]{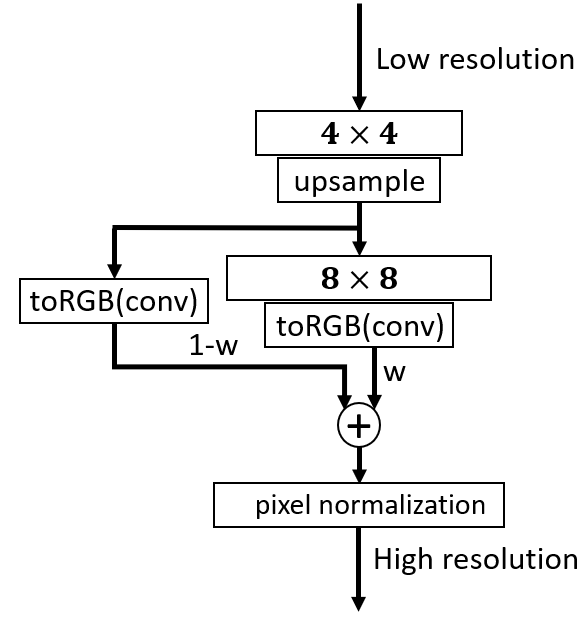}
\caption{The architecture of a layer in the PG-GAN.}
\label{fig:progan_architecture}
\end{wrapfigure}

In this work, we are studying what the disentanglement properties are in the ablation transformation of PG-GAN~(\cite{karras2017progressive}). PG-GAN is good at producing versatile objects with details while preserving the model efficiency. It adopts a progressive layer-growing strategy for fine-grained details and pixel normalization for training robustness.

Figure \ref{fig:progan_architecture} shows the architecture of a layer in PG-GAN which includes several different functions. Given a low-resolution input from upstream layers, we do upsampling, convolutions~(\cite{krizhevsky2012imagenet}), and weighted residual connection~\cite{he2016deep}. Finally, a pixel normalization~\cite{kurach2019large} which is a type of  avoids the competing gradient magnitudes spiraling out of control is imposed on the output as
\begin{gather}
b_{p,q} = \frac{a_{p,q}}{\beta_{p,q}} = \frac{a_{p,q}}{\gamma(a_{p,q})} = \frac{a_{p,q}}{\sqrt{\frac{1}{M}\sum_{j=0}^{M-1}(a_{p,q}^j)^2+\epsilon}} \label{eq:normalization}
\end{gather} 
where $M$ is the number of channels, $a_{p,q}$ and $b_{p,q}$ are the original and normalized feature vector at pixel $(p,q)$ and $\epsilon=10^{-8}$. And $\beta_{p,q}=a_{p,q}/\gamma(a_{p,q})$ refers to normalizing coefficients which implements the pixel normalization. 

The progressive growing strategy gradually appends new layers to the network, organizing layers into different granularity levels. Given a random input $\boldsymbol{z}$ and the PG-GAN generator $G=h\cdot f$, we have the generated image $\boldsymbol{x}=G(\boldsymbol{z})$ and $\boldsymbol{u}=f(\boldsymbol{z})$ is a set of units (i.e., channels in convolution operations) in a given middle layer which are closely related to classes of objects respectively. Literature shows that while units in the middle layers (layer4 to layer7 in PG-GAN) relate to object-level concepts (e.g. chairs, painting), they in layers ahead relate to background concepts (e.g. ceiling, sky) and in latter layers focus on abstract concepts (e.g. color, texture). In this work, to enable detailed analysis over objects' disentanglement properties, we constrain our training and following experiments of PG-GAN on the LSUN~(\cite{yu2015lsun}) conference room dataset.

PG-GAN's ability to yield various fine-grained objects is critical for analyzing disentanglement between different objects. In the literature, researchers conduct a wide range of experiments to primarily show that PG-GAN exhibits both disentangled and entangled properties in different scenarios. Especially in~(\cite{bau2018gan}), the authors present a series of qualitative unit-level experiments to show that in the unit-ablation transformation, while in some cases the PG-GAN shows disentanglement (such as ablating paintings on the wall), in other cases, entangled phenomena are observed, mainly categorized into two types including \textit{unsuppressible} and \textit{emerged} objects after ablation. In this work, we try to provide deeper thoughts about the disentanglement caused by function-level and mutual relationships between objects and eventually explain the above unexpected entanglement.

\section{Problem Definition} \label{sec:problem}
Intuitively, the disentanglement of deep models denotes the scenario that a transformation operating on a local component does not disturb other components in the same figure. A rigorous definition of transformation disentanglement is proposed by~\cite{besserve2018counterfactuals}, which is presented as below:

\begin{citeDef}[Counterfactual-based Disentanglement]
    Given a transformation $T$ on the data manifold, it is disentangled on a generative model $G$ with respect to a subset $\mathcal{E}$ of the generated outcome, if there is a transformation $T^\prime$ acting on internal representation units such that for any endogenous value $\boldsymbol{u}$
    \begin{gather}
        \boldsymbol{x}^\prime=T(G(\boldsymbol{u}))=G(T^\prime(\boldsymbol{u})),
    \end{gather}
    where $T^\prime$ only affects variables indexed by $\mathcal{E}$.
\end{citeDef}
This definition points out that a transformation $T$ is disentangled if and only if $T$'s effect corresponds to an internal transformation $T^\prime(\boldsymbol{u})$ which only causes changes on $\mathcal{E}$, a specific outcome subset. In this research, we specify $T$ in the above definition as the unit-ablation transformation, which is denoted by $T_a$. Therefore, $T_a$ denotes the transformation directly ablating objects, while $T^\prime_a$ is the transformation ablating input units. We further define $\mathcal{E}$ to the area of a specific object class $\mathcal{E}_c$ on the generated image. We also notice that $T_a$ is disentangled if and only if there is $T^\prime_a$ disentangled. Therefore, in the rest of the paper, we use ``disentanglement'' for short to represent the disentangled property of the ablation transformation $T^\prime_a$.

The above definition allows us to examine whether a type of objects are disentangled under the unit-ablation transformation on PG-GAN generated figures. For example, given a GAN $G$, if the $T_a$ on chair objects area $\mathcal{E}_c$ is disentangled, we expect to find a internal transformation $T^\prime_a$ acting on units $\boldsymbol{u}$ leading to only and sufficient disappearance of chair objects in the generated image. Therefore, we have the following definition of the disentanglement of a class of objectives. 
\begin{myDef}[The Disentanglement of a Set of Objects]
A class of objectives $\mathcal{C}$ is disentangled under the ablation transformation $T_a$, if the unit-ablation transformation $T^\prime_a$ acting on $\boldsymbol{u}$ satisfies
    \begin{gather}
        \boldsymbol{x}^\prime=T_a(G(\boldsymbol{u}))=G(T^\prime_a(\boldsymbol{u})),
    \end{gather}
    where $T_a^\prime$ only affects variables indexed by $\mathcal{C}$.
\end{myDef}

The above definition also suggest that the disentanglement of an object is influenced by several factors, among which the features of units $\boldsymbol{u}$ have the most direct and apparent effect because $T^\prime_a$ directly acts on $\boldsymbol{u}$. To quantitatively identify units $\boldsymbol{u}$'s impact on the appearance of a specific class of objects, following~(\cite{bau2018gan}) and \citep{holland1988causal}, we leverage a standard causal metric---the average causal effect (ACE) to reflect a unit's effect on disentanglement. 

\begin{myDef}[Unit's Effect on Disentanglement]
    For any possible $\boldsymbol{z}$ and $\boldsymbol{x}=G(\boldsymbol{z})$, the ACE of unit set $U\in\boldsymbol{u}$ on object class $c$ is defined as
\begin{gather}
\delta_{U\rightarrow c}=\mathbb{E}_{\boldsymbol{z}}[\boldsymbol{S}_c(\boldsymbol{x}_i)]-\mathbb{E}_{\boldsymbol{z}}[\boldsymbol{S}_c(\boldsymbol{x}_a)] \label{equ:gan-dissection}
\end{gather}
where $\boldsymbol{x}_a=h(T^\prime_a(\boldsymbol{u}_{U}))=f(0,\boldsymbol{u}_{\overline{U}})$ is the image with $U$ ablated at $P$ and $\boldsymbol{x}_i=f(k,\boldsymbol{u}_{\overline{U}})$ is the image with $U$ inserted at $P$.
\end{myDef}

\begin{figure}[t]
    \centering
    \begin{subfigure}[t]{1.8in}
    \includegraphics[width=1.8in]{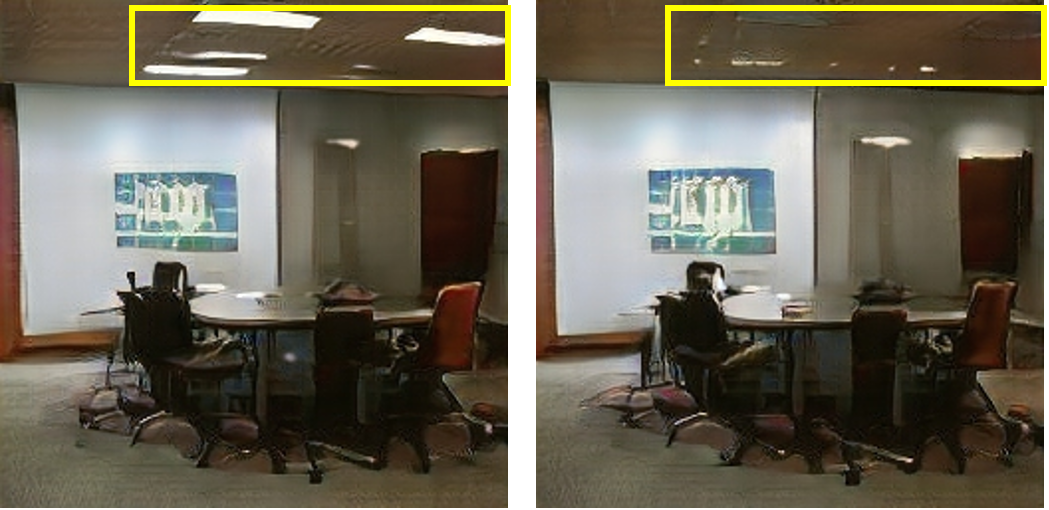}
    \caption{ablate light units}
    \end{subfigure}
    \begin{subfigure}[t]{1.8in}
    \includegraphics[width=1.8in]{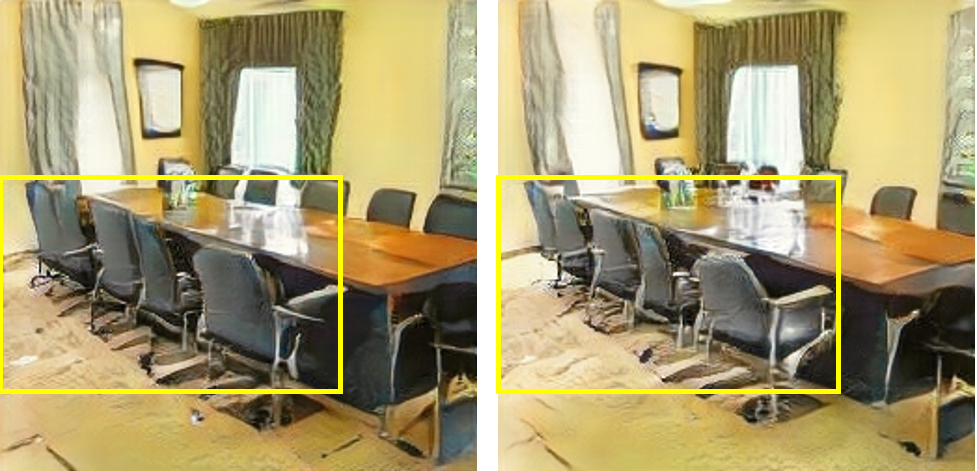}
    \caption{ablate chair units}
    \end{subfigure}
    \begin{subfigure}[t]{1.8in}
    \includegraphics[width=1.8in]{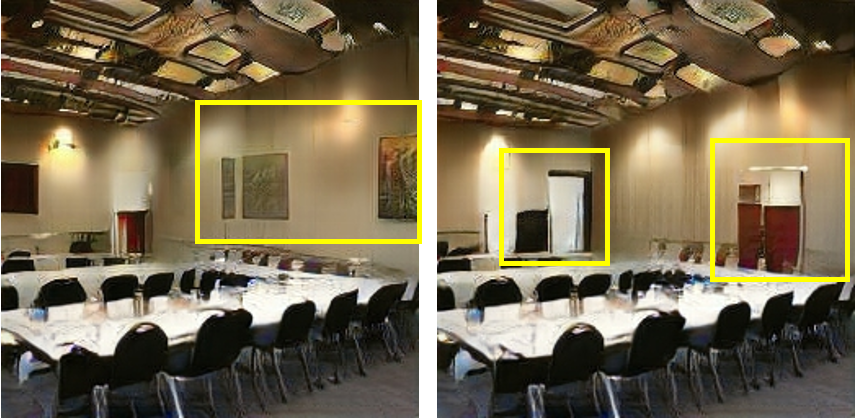}
    \caption{ablate painting units}
    \end{subfigure}
    \caption{(a) Disentangled case: ablating top-20 light-ACE units eliminates the light objects on ceiling. (b) Deactivating-ablation entanglement: ablating chair-ACE units cannot eliminate the chairs. (c) Mis-ablation entanglement/emerge case: ablating top-20 painting-ACE units leads to both elimination of paintings on the wall and unexpected emergence of doors and windows}
    \label{fig:problem}
\end{figure}

Note that the $\delta_{U\rightarrow c}$ is the expectation over all the possible $\boldsymbol{z}$, which reflects the average effect of units on the object class $c$ in any scenarios. In Figure \ref{fig:problem}, we exhibit the three types of disentangled and entangled phenomena caused by ablating units $U$ with top-$\delta_{U\rightarrow c}$. The first sample shows a fully disentangled behavior (as the paintings on the wall disappear), but the second and third exhibit entangled phenomena. In the second one, although top-$\delta$ chair units are ablated, the chair objects are \textit{not ablated} in the outcome, with only sizes shrinking a little bit. In the third one, while paintings are removed, new objects such as doors unexpectedly \textit{emerge}.
 
Literature has suggested that in the internal components of neural networks may have roles in causing entanglement \citep{besserve2018intrinsic,chen2018isolating}. In this work, we propose to study the pixel normalization function's effect on disentanglement $T^\prime_a$ via experimental designs. We define it as follows:

\begin{myDef}[Pixel Normalization's Effect on Disentanglement]
    Given an input $\boldsymbol{z}$, let $\boldsymbol{u}^\prime=T^\prime_a(\boldsymbol{u})$ denote ablated units, $\varphi(\beta,\boldsymbol{u})$ denote pixel normalization function's effect, and $\theta(\boldsymbol{u})$ denote other functions' effect, there is
    \begin{gather}
        Y(\beta,\boldsymbol{u})=\varphi(\beta,\boldsymbol{u}) + \theta(\boldsymbol{u})
        \label{equ:Y}
    \end{gather}
    where $Y$ refers to the functions and units' joint effect on $\boldsymbol{x}$. Consequently, the effect of pixel normalization given the ablation transformation $T^\prime_a$ could be represented by
    \begin{gather} \Delta Y_{\beta\to\beta^\prime} \text{,\ where\ }\beta^\prime=\gamma(\boldsymbol{u}^\prime)=\gamma(T^\prime_a(\boldsymbol{u}))=T^\prime_a(\gamma(\boldsymbol{u}))=T^\prime_a(\beta)
    \end{gather}
\end{myDef}

Since the pixel normalization acts on the units, $\varphi$'s behavior is probably influenced by the aggregating properties of the input $\boldsymbol{u}$ and corresponding $\delta_{\boldsymbol{u}\rightarrow c}$. More specifically, 

\begin{myDef}[Distribution and Ranking (Informal)]
    With a set of object classes $\mathcal{C}$ and the set of units $\boldsymbol{u}$, we informally define:
    
    1) \textbf{distribution of} $\delta_{u\rightarrow c}$: given an object class $c\in\mathcal{C}$ and for every unit $u_i\in\boldsymbol{u}$, the density distribution of $\delta_{u_i\rightarrow c}$
    
    2) \textbf{ranking of} $\delta_{u\rightarrow c}$: for every $c\in\mathcal{C}$, the ranking sequence of their $\sum_i \delta_{u_i\rightarrow c}$
\end{myDef}
 
\hide{
\begin{enumerate}
    \item All units causally contribute to more than one object class respectively. Some units have a dominant impact on only one object class, and others could have strong contributions to multiple classes.\xiao{yu: incorrect arrangement}
    \item Given a class of objects to remove, it is usually sufficient to make it by ablating units with dominant influence. But we also observe entangled phenomena including \textit{unsuppressible} (minimal effect on $\mathcal{E}$) and \textit{emerged} (significant effect beyond $\mathcal{E}$) objects after ablation.
\end{enumerate}
\xiao{from definition induces why this is Figure 2 disentangled. no (a) (b), leave (c)(d)(e)}
}

\section{Method: Difference-in-Difference experiment design}
To bridge the pixel normalization to disentanglement, in Section \ref{sec:problem} we propose to identify $\Delta Y_{\beta\to\beta^\prime}$ which represents pixel normalization's universal effect, and then to explore how the properties of $\boldsymbol{u}$ will influence $\Delta Y_{\beta\to\beta^\prime}$'s form of expression. For the first problem, we introduce a counterfactual-based Difference-in-Difference (DID) \citep{2018Dif} experiment framework; for the second, based on the different-in-different approach, we compare $\Delta Y_{\beta\to\beta^\prime}$ under four scenarios with different unit distributions and rankings.

\subsection{Counterfactual DID experiments for identifying function's effect} \label{sec:beta-design}
We argue that the pixel normalization's transformation $\beta\rightarrow\beta^\prime$ has a significant impact on disentanglement of the $\boldsymbol{u}\rightarrow\boldsymbol{u}^\prime$. The intuition is that, from Equation \ref{eq:normalization}, we know that $\beta_{p,q}$ will increase when there are units ablated (i.e. $a^j_{p,q}=0$), leading to augmentations on un-ablated units. In previous study, researcher show the object-level generation of PG-GAN is determined by the 4-layer to 13-layer in PG-GAN. Therefore, in order to investigate how the $\beta\rightarrow\beta^\prime$ affects the transformation $\boldsymbol{u}\rightarrow\boldsymbol{u}^\prime$, we propose to control the $\beta$ in these 10 layers for constructing counterfactuals.

Naturally, we have a factual experiment $Y(\beta,\boldsymbol{u})$ as shown in Equation \ref{eq:exp_design}. To show $\beta$'s effect on the transformation $\boldsymbol{u}\rightarrow\boldsymbol{u}^\prime$, we construct a counterfactual case where $\boldsymbol{u}\rightarrow\boldsymbol{u}^\prime$ and $\beta$ remains controlled. Technically speaking, this control is performed by enforcing normalizing coefficients $\beta$ in 4-layer to 13-layer to be the same as the original situation with $\boldsymbol{u}$. For another factual experiment $Y(\beta^\prime,\boldsymbol{u}^\prime)$, we conduct the same control as enforcing coefficients to be the same to reveal $\beta^\prime$'s impact on $\boldsymbol{u}\rightarrow\boldsymbol{u}^\prime$. As a result, we construct two pairs of counterfactual experiments for difference in difference analysis as follow:
\begin{gather}\label{eq:exp_design}
    \text{Original case:}\qquad\qquad\qquad\qquad\qquad\qquad\qquad\ \  Y(\beta,\boldsymbol{u})=\varphi(\beta,\boldsymbol{u}) + \theta(\boldsymbol{u}) \\ 
    \text{Ablated~} \boldsymbol{u}^\prime \text{~and~} \beta\text{:}\qquad\qquad\qquad\qquad\qquad\quad\quad\ Y(\beta,\boldsymbol{u}^\prime)=\varphi(\beta,\boldsymbol{u}^\prime) + \theta(\boldsymbol{u}^\prime) \notag\\
    \text{Ablated~} \boldsymbol{u}^\prime \text{~and~} \beta^\prime \text{:\qquad\qquad\qquad\qquad\qquad\quad\ \ } Y(\beta^\prime,\boldsymbol{u}^\prime)=\varphi(\beta^\prime,\boldsymbol{u}^\prime) + \theta(\boldsymbol{u}^\prime) \notag\\
    \boldsymbol{u} \text{~and ablated~} \beta^\prime\text{:}\qquad\qquad\qquad\qquad\qquad\qquad\quad Y(\beta^\prime,\boldsymbol{u})=\varphi(\beta^\prime,\boldsymbol{u}) + \theta(\boldsymbol{u}) \notag
\end{gather} 
and therefore the real effect could be computed as:
\begin{align}
    \Delta Y_{\beta\to\beta^\prime}
    &=[Y(\beta^\prime,\boldsymbol{u}^\prime)-Y(\beta^\prime,\boldsymbol{u})]-[Y(\beta,\boldsymbol{u}^\prime)-Y(\beta,\boldsymbol{u})]\\
    &=[\varphi(\beta^\prime,\boldsymbol{u}^\prime)-\varphi(\beta^\prime,\boldsymbol{u})] -[\varphi(\beta,\boldsymbol{u}^\prime)-\varphi(\beta,\boldsymbol{u})]+ [\theta(\boldsymbol{u})-\theta(\boldsymbol{u})] + [\theta(\boldsymbol{u}^\prime)-\theta(\boldsymbol{u}^\prime)]\notag\\
    &=[\varphi(\beta,\boldsymbol{u}^\prime)-\varphi(\beta,\boldsymbol{u})]-[\varphi(\beta^\prime,\boldsymbol{u}^\prime)-\varphi(\beta^\prime,\boldsymbol{u})]\notag
\end{align}
which indicates the real effect of pixel normalization effect function $\varphi$ imposed on the image. Noted that this effect is also unbiased because the unrelated effect $\theta$ has been offset in the difference.

\subsection{Scenario-conditioned experiments for unit distribution and ranking}
Difference-in-Difference experiments enable us to systematically analyze pixel normalization's effects on disentanglement of any given objects on given images. But we may still not explain the unexpected evidences in Figure \ref{fig:problem}, that with the same application of pixel normalization, different object classes, or the same object class in different scenarios show significant entangled properties.

Therefore, in this step of experiment, we design to compare the DID experiment results conditioned on its distribution of $\delta_{u\rightarrow c}$ and ranking of $\delta_{u\rightarrow c}$. More specifically, we categorized them into $2\times2=4$ types: 
\begin{enumerate}
    \item distribution of $\delta_{u\rightarrow c}$: 1) units symmetrically locate in both high-ACE and low-ACE area, 2) units concentrate in high-ACE area.
    \item ranking of $\delta_{u\rightarrow c}$: 1) the top ranking object's ACE overwhelms the second one, 2) the top ranking object marginally surpasses the second one.
\end{enumerate}

Although theoretically we have 4 scenarios, in fact we will show that given the symmetrical distribution condition, the ranking of $\delta_{u\rightarrow c}$ does not take effect, leaving only 3 meaningful cases.

\section{Result}
\subsection{Pixel normalization's effect: entangle objects by in-painting} \label{sec:beta-result} 
Our experiments demonstrate a strong tie between pixel normalization and the entanglement in the object ablation transformation $T^\prime_a$. Supported by difference-in-difference counterfactual experiments, we discover that the pixel normalization's effect $\Delta Y_{\beta\rightarrow\beta^\prime}$, informally, is to in-paint objects to ablated areas on the generated image caused by the internal ablation of corresponding units, which consequently leads to entanglement in PG-GAN. In Figure 4, we present the mechanism of how pixel normalization causes entanglement. Recall the difference-in-difference definition of pixel normalization's effect as:
\begin{gather}
    \Delta Y_{\beta\rightarrow\beta^\prime}=[Y(\beta^\prime,\boldsymbol{u}^\prime)-Y(\beta^\prime,\boldsymbol{u})]-[Y(\beta,\boldsymbol{u}^\prime)-Y(\beta,\boldsymbol{u})]
\end{gather}
of which the first term $Y(\beta^\prime,\boldsymbol{u}^\prime)-Y(\beta^\prime,\boldsymbol{u})$ describes the in-painting effect produced by $\beta^\prime$ (first two columns of Row 1) and the second term $Y(\beta,\boldsymbol{u})-Y(\beta,\boldsymbol{u}^\prime)$ depicts an ablation effect (first two columns of Row 2). As a consequence of their difference, the in-painted area will cover the ablation area and presents a joint effect of in-painting. In order to well present the in-painting effect and ablation effect, we present the two effects on a background figure $\mathcal{B}$. The background figure presents the areas that do not significantly change by both effects. The background figure is presented on the third column of Figure 4. 


\vpara{In-painting effect under $\beta^\prime$ (after ablation)} When $\beta^\prime$ is controlled, from $Y(\beta^\prime, \boldsymbol{u})$ to its counterfactual $Y(\beta^\prime, \boldsymbol{u}^\prime)$, pixel normalization shows its in-painting functionality by inserting door objects on the original wall and wall objects on where the paintings are.

\vpara{Ablation effect under $\beta$ (before ablation)} When $\beta$ is controlled, pixel normalization's response is to ablate the area of all objects related to the removed units. For example, in Figure 4's lower row where we ablated units mainly contribute to the paintings, from the original case $Y(\beta,\boldsymbol{u})$ to the counterfactual case $Y(\beta,  \boldsymbol{u}^\prime)$, the PG-GAN removes all paintings away together with surrounding wall areas and turning them into a yellow background.

\vpara{Joint DID effect of the pixel normalization} Combining the above two effects of in-painting and ablation, we present the final DID joint effect as in the rightmost subfigure in Figure 4, which clearly illustrates that the pixel normalization in-paints not only walls at where the paintings are ablated, unexpected door objects are also inserted to places where should have been wall. Without the rigorous DID experimental design, one may only notice the effect from the original case $Y(\beta, \boldsymbol{u})$ to the natural ablation case $Y(\beta^\prime, \boldsymbol{u}^\prime)$, and thus ignore the complicated mechanisms behind the pixel normalization. 

\begin{figure}[t] \label{fig:equation}
    \centering
    \includegraphics[width=5.2in]{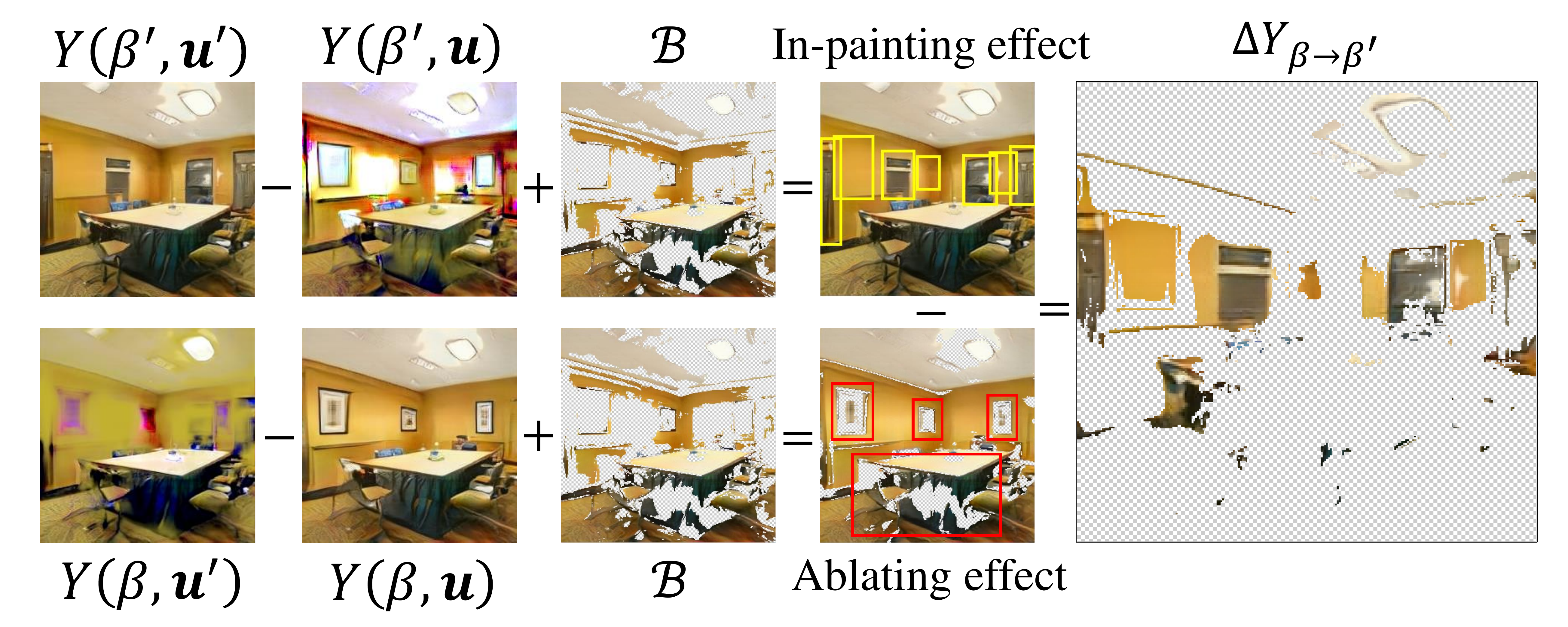}
    \caption{The mechanism of how pixel normalization causes entanglement. The first row represents the mechanism of how ablated figure is changed by adding the top 20 related units back when pixel normalization coefficient is fixed to the after-ablation level. The second row represents the mechanism of how original figure is changed by ablating the top 20 related units back when pixel normalization coefficient is fixed to the before-ablation level. The right large figure represents the effect of pixel normalization.
    \vspace{-0.7cm}
}
\end{figure}

\subsection{What to in-paint: unit distribution and ranking}
In Section \ref{sec:beta-result}, we have answered the universal effect of pixel normalization, that to in-paint the area affected by the ablation transformation. But what objects the pixel normalization will in-paint to the area and how intensive this in-painting will remain unclear. We argue that the form of in-painting the pixel normalization will present in an ablation transformation could be identified according to the properties of units' ACE $\delta_{u\rightarrow c}$, namely the distribution and the ranking.

First, the distribution of $\delta_{u\rightarrow c}$ determines whether an object class could be ablated. In Figure 5, we select three typical types of object classes, including chair, painting and light, to illustrate how the ACE distribution directly determines an object's ability to be disentangled. We ablate the units with ACE ranking from top to down, and observe an apparent gap between chair (categorized as Type-1) objects and painting \& light (categorized as Type -2 \& disentangled). While the chairs' area remains stubborn with top-80 ACE units ablated, the painting and light objects take no more than 20 of the top ACE units ablated to be eliminated. 

This confusing phenomenon could be explained when we look into their distribution of ACE. We observe that Type-1 classes distribution is flat, with high-ACE units explains for no more than 50\% of the total effect. Therefore, when we removing high-ACE units, the loss of their effects could be compensated by those lower-ACE units with $\beta^\prime$'s augmentation. However, on the contrary, Type-2 and disentangled classes have a concentrated distribution where influential units dominantly locate at the high-ACE part and thus make the most contributions to the object generation. To conclude, with such flat distribution, Type-1 classes are not disentangled at all with the ablation transformation.

\begin{figure}[t]
\centering
\begin{minipage}[t]{0.55\textwidth}
\centering
\includegraphics[width=7.5cm]{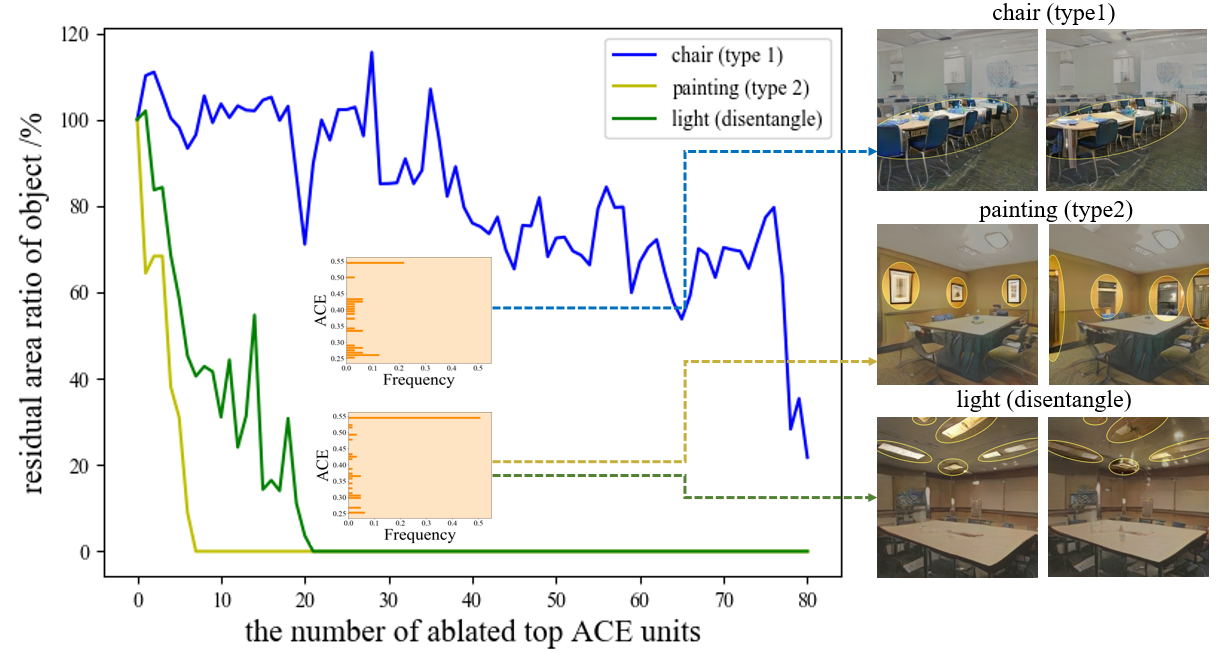}
\setcaptionwidth{2.8in}
\caption{ACE Distribution determines whether and how entanglement occurs. }

\label{segarea-3typre}
\end{minipage}
\begin{minipage}[t]{0.35\textwidth}
\centering
\includegraphics[width=5.0cm]{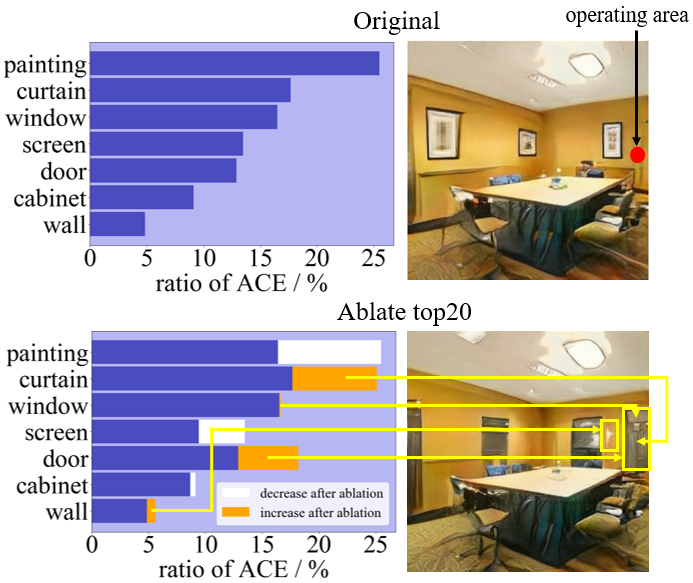}
\setcaptionwidth{2.0in}
\setlength{\abovecaptionskip}{0cm}
\caption{ The changes in ACE ranking and ratio determine what types objective are used to in-painted}


\label{33-ranking}
\end{minipage}
\end{figure}

Second, given the Type-2 and disentangled object classes, ablation on corresponding units could eliminate them. But for Type-2 classes, when we compare it with those disentangled ones, we observe mis-ablation phenomena in which other unexpected objects would emerge. To explain this, based on the concentrated unit distribution we consider the influence of the ranking of units as shown in Figure \ref{33-ranking}. When we operate ablation on the red point and examine its unit ranking, before ablation ACE for painting is the largest, but ACE for window, curtain and door are also quite large. After ablation, the decrease of painting ACE and the increase of wall ACE make painting removed and replaced with wall. Moreover, ACE for curtain, door and window become the largest and they indeed appear in the area around the operation point, which successfully explains the mis-ablation.

For the disentangled classes, such as the lights we display in the summary plot Figure 1, they actually share a similar ranking of units with those of Type-2. The only difference is that the second candidate unit just behind the ablated one happens to be the original surrounding.

\section{Conclusion}We in this research propose a difference-in-difference (DID) counterfactual framework to design experiments for acquiring insights into the black box of PG-GAN transformation and analyzing the entanglement mechanism in one of the Progressive-growing GAN  (PG-GAN). Our experiment clarifies the mechanism of how pixel normalization causes PG-GAN entanglement during an input-unit-ablation transformation. We discover that pixel normalization causes object entanglement by in-painting the area occupied by ablated objects. We also discover the unit-object relation determines whether and how pixel normalization causes object entanglement. Our DID framework theoretically guarantees that the mechanisms that we discover is solid, explainable, and comprehensive.


\bibliography{iclr2021_conference}
\bibliographystyle{iclr2021_conference}

\end{document}